\documentclass[lettersize,journal]{IEEEtran}
\usepackage{amsmath,amsfonts}
\usepackage{algorithmic}
\usepackage{algorithm}
\usepackage{array}
\usepackage{authblk}
\usepackage[caption=false,font=normalsize,labelfont=sf,textfont=sf]{subfig}
\usepackage{textcomp}
\usepackage{stfloats}
\usepackage{url}
\usepackage{verbatim}
\usepackage{graphicx}
\usepackage{cite}
\usepackage{booktabs} 
\usepackage{hyperref}

\usepackage{tabularx}
\usepackage{amsfonts}
\usepackage{color}
\usepackage{array}

\usepackage{graphicx} 
\graphicspath{ {./figures/} }

\newcommand{\by}{\mathbf{y}}

\newcommand{\bY}{\mathbf{Y}}

\newcommand{\bx}{\mathbf{x}}
\newcommand{\bw}{\mathbf{w}}

\newcommand{\btheta}{\boldsymbol{\theta}}
\newcommand{\bTheta}{\boldsymbol{\Theta}}

\newcommand{\ourmethodlongit}{\textit{Distributional Latent Variable Modeling}}

\newcommand{\ourmethod}{\texttt{DLVM}}
\newcommand{\ouractivemethod}{\texttt{DALE}}
\newcommand{\oldmethod}{\texttt{IMLE}}
\newcommand{\train}{\texttt{COLL10}}
\newcommand{\validref}{\texttt{TB}}
\newcommand{\validnew}{\texttt{ML}}
\newcommand{\reliab}{\texttt{MLR}}
\newcommand{\modelref}{\texttt{TB-IMLE}}
\newcommand{\modelnew}{\texttt{ML-DLVM}}

\begin{document}

\title{Distributional Latent Variable Models with an Application in Active Cognitive Testing}


\author{
    Robert Kasumba\thanks{Email: rkasumba@wustl.edu}, 
    Dom C.P. Marticorena\thanks{Email: dom.m@wustl.edu}, 
    Anja Pahor\thanks{Email: a.pahor@northeastern.edu}, 
    Geetha Ramani\thanks{Email: gramani@umd.edu}, 
    Imani Goffney\thanks{Email: igoffney@umd.edu}, 
    Susanne M. Jaeggi\thanks{Email: smjaeggi@uci.edu}, 
    Aaron R. Seitz\thanks{Email: a.seitz@northeastern.edu}, 
    Jacob R. Gardner\thanks{Email: jacobrg@seas.upenn.edu}, 
    Dennis L. Barbour,~\IEEEmembership{Member,~IEEE}\thanks{Email: dbarbour@wustl.edu}
}


\renewcommand{\thefootnote}{\fnsymbol{footnote}}
\setcounter{footnote}{0}

\makeatletter
\renewcommand\@makefntext[1]{}
\makeatother



\maketitle

\begin{abstract}
Cognitive modeling commonly relies on asking participants to complete a battery of varied tests in order to estimate attention, working memory, and other latent variables. In many cases, these tests result in highly variable observation models. A near-ubiquitous approach is to repeat many observations for each test independently, resulting in a distribution over the outcomes from each test given to each subject. Latent variable models (LVMs), if employed, are only added after data collection. In this paper, we explore the usage of LVMs to enable learning across many correlated variables simultaneously. We extend LVMs to the setting where observed data for each subject are a series of observations from many different distributions, rather than simple vectors to be reconstructed. By embedding test battery results for individuals in a latent space that is trained jointly across a population, we can leverage correlations both between disparate test data for a single participant and between multiple participants. We then propose an active learning framework that leverages this model to conduct more efficient cognitive test batteries. We validate our approach by demonstrating with real-time data acquisition that it performs comparably to conventional methods in making item-level predictions with fewer test items.
\end{abstract}

\begin{IEEEkeywords}
Active machine learning, executive function, latent variable modeling, cognition.
\end{IEEEkeywords}

\IEEEpeerreviewmaketitle

\section{INTRODUCTION}
\label{sec:introduction}

%
%
%
%

\IEEEPARstart{M}{any} unobservable phenomena in the social and behavioral sciences are estimated by presenting test participants with repeated testing. For example, asking a participant to recall a sequence of independent items soon after studying them is one convenient and interpretable way to operationalize the latent cognitive construct of working memory \cite{wilhelm2013working}. Because this testing procedure is quite noisy, however, estimating a participant's working memory typically proceeds by repeatedly querying related test items many times and averaging the results \cite{schmiedek2014task}. This procedure may be tractable for one-dimensional constructs such as working memory, but probing more interesting complex scenarios is less so.

For example, tracking overall cognitive function is crucial for those at risk of dementia \cite{weintraub2018version}, while assessing the cognitive readiness of students just before a math lesson can significantly impact educational outcomes \cite{lee2009contributions,bull2011using}. In such cases, cognitive meta constructs such as executive functions (EFs) play a pivotal role \cite{jurado2007elusive}. Typically, test batteries are employed to explore these phenomena, incorporating multiple tests for each underlying construct to ensure the results are generalizable \cite{akshoomoff2014nih,weintraub2013cognition}. The extensive data and time requirements for conducting a full battery of cognitive tests can be prohibitive, however, hindering widespread implementation \cite{rouder2023correlations, rouder2024cronbach}.

Moreover, while standard methods like independent Maximum Likelihood Estimation (MLE) can estimate key metrics such as mean reaction times, they fall short in capturing the uncertainty in underlying cognitive behavior and the measurement process. Advanced methods such as structural equation modeling (SEM), which attempt to extract cross-information from different cognitive tests, often assume linear interactions between tests—an assumption that may not always hold true. They are also impractical to model random effects at the individual level. 

To overcome these significant challenges, we propose a novel approach: non-linear \ourmethodlongit{} (\ourmethod{}). This innovative method not only leverages the similarities and cross-information across different individuals to provide precise estimates of metrics of interest but also quantifies the process uncertainty, offering a more robust and reliable understanding of cognitive performance.

We formalize our approach using following broader modeling problem statement. We assume that each of $n$ participants is given a battery of $T$ tests. The $i$th participant taking the $t$th test performs $s_{ij}$ repetitions of that test, resulting in a series of observations $y_{it}^{(1)},...,y_{it}^{(s_{it})}$. We seek to develop a joint model of all observations $Y = \left[\by_{it}\right]_{i=1..n,t=1..T}$ that will enable us to fully capture a participant's results on all $T$ tests in the battery while collecting significantly fewer than the full $\sum_{t}s_{it}$ trials (i.e., test items) of the typical test battery described above. To accomplish this, we will leverage correlations both across different tests in the battery for a given individual, and across a population of tested individuals.

In contrast to the set of high dimensional vectors $Y$ in the traditional LVM setting \cite{bayesGPLVM}, for \ourmethod{} we are given data $\by_{it}$ assumed to be drawn from a heterogeneous set of distributions $p(\by_{it} \mid \btheta_{it})$. We develop a Bayesian hierarchical model in this setting as well as an accompanying variational inference procedure. 


Beyond the \ourmethod{} model, we develop a Bayesian active learning procedure \cite{chaloner1995bayesian,cohn1996active}. This procedure selects trials in the test battery sequentially by maximizing the mutual information (e.g., \cite{gardner2015psychophysical} for use of an analytical approximation for psychometric testing) between the test to be performed and the latent variables for a new participant. 


We trained the \ourmethod{} model using a real-world dataset of cognitive test battery data collected from 18 young adults over 10 separate test sessions. We validated our approach by demonstrating accuracy, efficiency and reliability on two additional groups of young adults in new experiments, showing that \ourmethod{} performs comparably to conventional methods in the accuracy item-level predictions, but more efficiently with fewer test items. 

In summary, we make the following key contributions in this work.
\begin{itemize}
    \item We introduce a non-linear \ourmethodlongit{} capable of quantifying process and model uncertainty on top of estimates of the main cognitive metrics provided by traditional methods.
    \item Using \ourmethod{} we demonstrate modeling individuals' cognitive test performance at the item level, offering a more granular understanding of their cognitive abilities.
    \item We introduce an active learning framework that leverages \ourmethod{} to identify the most informative cognitive tests to deliver leading to fewer tests needed to represent cognitive performance.
\end{itemize}

\section{BACKGROUND AND RELATED WORK}
\label{sec: background}
\begin{figure*}[!ht]
\includegraphics[width=\textwidth]{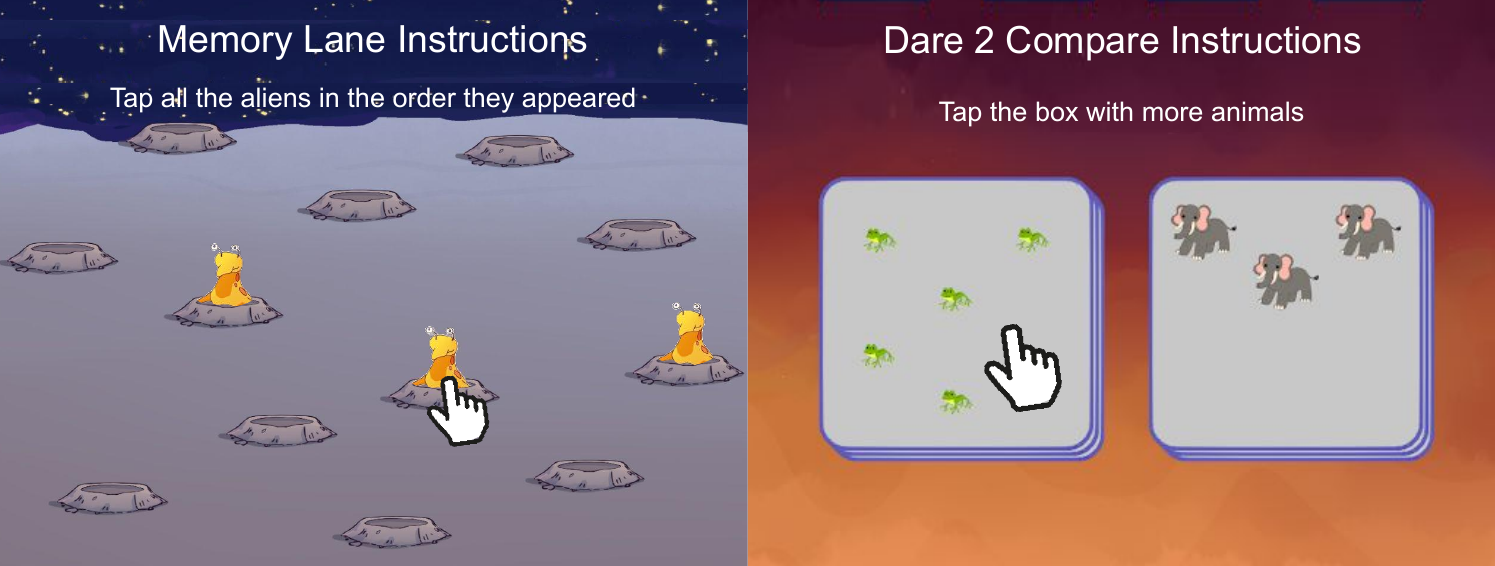}
    \caption{Two example tests in the battery. Left: Span---Recall the order aliens appeared (binary accuracy). Right: Numerical Stroop---Select the box with more animals (reaction time).}
    \label{fig:testbattery}
\end{figure*}
\subsection{Cognitive test batteries}
Inference over cognitive constructs typically begins with behavioral data acquired from participants completing a series of independent tests arranged in a battery \cite{weintraub2013cognition}. The specific cognitive variables evaluated in this paper are the executive functions (EFs) of working memory, inhibitory control, and cognitive flexibility \cite{jurado2007elusive}. Altogether we employ 7 tests with output distributions parameterized by a total of $D=12$ parameters, each targeting one of these 3 EFs. Examples of the tests can be seen in \autoref{fig:testbattery}. Tests were designed in appearance and difficulty for children of ages $\sim$10–12 because our primary interest is efficiently tracking their EFs in order to optimize the design of their individual math lessons.

Repeated iterations of these tests generate trials that are best modeled by different distributions. In the span task, for example, the user is presented with a sequence of $k$ aliens and must tap them in the correct order. As $k$ varies, we model the probability that the participant will recall the correct sequence as a sigmoidal function of span length, $\sigma\left(\frac{k + \theta_\psi}{\sigma_\psi}\right)$. By collecting repeated trials for various $k$, we estimate psychometric threshold $\theta_\psi$ and spread $\sigma_\psi$. The numerical Stroop timing task, on the other hand, requires participants to tap as quickly as possible the box with more animals displayed. Most responses here are correct, so the variable of interest is the participant's reaction time, which we model with a log-normal distribution $\log \mathcal{N}(\mu_{\tau}, \sigma_{\tau}^2)$. By collecting repeated samples from trials of this test, we estimate reaction time mean $\mu_{\tau}$ and variance $\sigma_{\tau}^2$ for each participant.

\subsection{Modeling cognitive performance}
The conventional approach handles one cognitive test at a time, each assumed to target primarily a specific executive function (EF) construct. Due to noise in the observations caused by either measurement errors or the underlying cognitive process, repeated trials are delivered, and Maximum Likelihood Estimation (MLE) is used to estimate the individual's performance\cite{schmiedek2014task,embretson1991multidimensional,cousineau2004fitting}. MLE is a robust method for parameter estimation that finds the parameter values that make the observed data most probable. This approach requires a substantial amount of data from an individual for all tests delivered, however \cite{rouder2024cronbach,rouder2024hierarchical, rouder2023correlations}, which can be resource-intensive. Additionally, it assumes that data from an individual's performance on one cognitive test is not informative about their performance on other tests. This independence assumption limits the potential to leverage shared information across different tests.

\subsection{Latent variable models (LVMs)}

Our approach is closely related to the Latent Variable Model (LVM) framework. LVMs are sophisticated mathematical methods designed to infer unobserved variables from directly measured values \cite{bartholomew2011latent}. These models are essential for reducing dimensionality in multivariate analysis, playing a pivotal role in uncovering the hidden factors that influence observed outcomes. In cognitive studies, LVMs have traditionally utilized factor analysis techniques, specifically exploratory factor analysis (EFA) and confirmatory factor analysis (CFA) \cite{vandekerckhove2014cognitive}. These methods presuppose linear relationships between observed cognitive performances and their underlying latent structures.

Structural Equation Modeling (SEM) extends these factor analysis techniques by incorporating a predefined structure for the interactions between cognitive test performances as well as the latent variables \cite{kline2023principles, friedman2009individual}. This widely employed method not only assumes that the relationship between test-level performances is linear but also requires a predefined structure which may not always hold in complex interactions such as those related to executive functioning.

Another LVM-based approach is the autoencoder framework, which takes a feature space vector, projects it to a low-dimensional latent space, and then reconstructs the feature space vector from this latent space. These autoencoders, often modeled using neural networks, are capable of capturing non-linear relationships \cite{aggarwal2018neural}. Variational autoencoders (VAEs) \cite{kingma2019introduction} are a probabilistic variation that aims to regularize the construction of the latent space. For example, \cite{xue2022perioperative} constrained the latent variables to be predictive of outcomes, as well.

These LVM approaches typically require a high dimensional input vector \cite{bayesGPLVM} which in our case would be performance across different cognitive tests, making them unable to model individual-level random effects in performances across tests. This would be possible if they could fit an individual likelihood for each test. In contrast, our generalized approach can model cognitive test performance at the item level, thereby capturing random effects. This capability allows for a more detailed and individualized understanding of cognitive performance.

\section{METHODS}
\label{sec:methods}

We assume we have a collection of $n$ training participants, each of which has taken a battery of $T$ cognitive tests. For participant $i$, taking a cognitive test $t$ involves collecting $s$ observations via repeated trials of that test $y_{it}^{(1)},...,y_{it}^{(s_{it})}$. The observations $y_{it}$ have different forms across the various cognitive tests. For example, a test measuring working memory on $k$ items may have binary observations of whether or not the participant successfully recalled all of those items, while a test measuring reaction time in response to a particular cognitive task may have positive real-valued observations. Furthermore, any parameters of distributions used to model test responses must not be shared directly across participants, as individuals may obviously have different performance on the same cognitive test (e.g., random effects).
\begin{figure*}[!ht]
\centering
\includegraphics[width=0.9\textwidth]{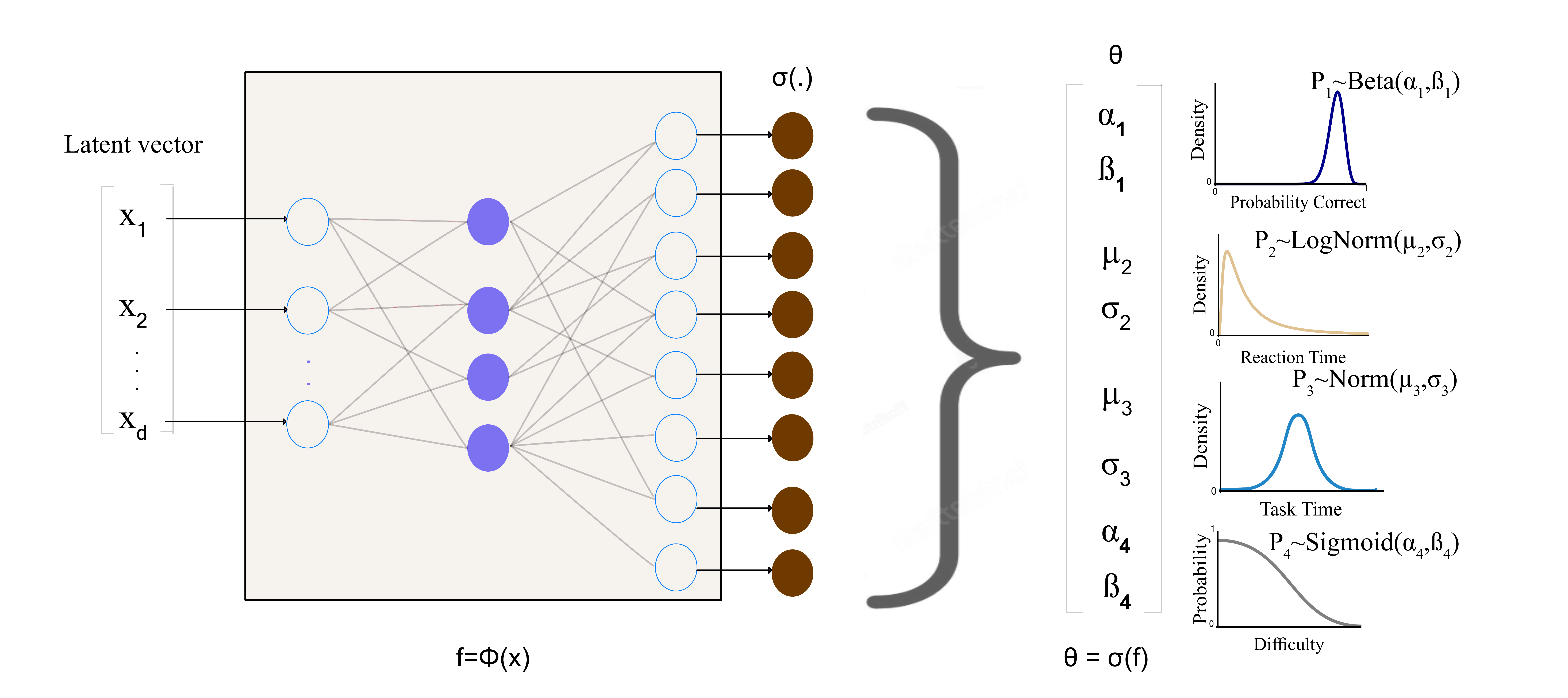}
    \caption{\ourmethod{} model architecture. The model only utilizes the decoder component of the traditional autoencoder framework. Given a latent vector \( x \), it outputs the distributional parameters \( \theta \). The model is trained to maximize the likelihood of the observed item-level data \( Y_i \) under the output distributional parameters $\theta_i$.
}
    \label{fig:dlvm-architecture}
\end{figure*}

\subsection{Latent variable models for distributional data}
Because multiple tests are typically given to measure the same cognitive construct (e.g., working memory), and because there is strong reason to believe that many of the tests may estimate multiple cognitive constructs of interest \cite{friedman2017unity}, a primary challenge in this setting is developing a model that enables us to leverage correlations (1) across the various cognitive tests given to a single individual, and (2) across individuals in the training population.

\paragraph{Observation model} We model the observations for an individual $i$ and a specific cognitive test $t$, $\by_{it}$, as being generated from distributions $p(\by_{it} \mid \btheta_{it})$. We assume the observations for cognitive test $p$ and cognitive test $q$ are conditionally independent given the parameters $\btheta_{ip}$ and $\btheta_{iq}$. Denote by $\bTheta_{i} = \left\{\btheta_{i1},...,\btheta_{iT}\right\}$ the set of $D$ parameters across all cognitive tests for participant $i$. This results in the joint distribution over the full set of observations $\by_{i}$ for participant $i$ to be given by:
\begin{equation*}
    p(\by_{i} \mid \bTheta_{i}) = \prod_{t=1}^{T} p(\by_{it} \mid \btheta_{it})
\end{equation*}
\paragraph{Latent variable model}  We model the full set of $\bTheta \in \mathbb{R}^{n \times D}$ parameters for all participants as being associated with $d$ latent variables $X \in \mathbb{R}^{n \times d}$, with $d \leq D$, as is common in, e.g., variational autoencoders \cite{kingma2019introduction}. We place independent standard normal priors over all latent variables:
\begin{equation*}
    p(X) = \mathcal{N}(0, \mathbf{I}_{nd}),
\end{equation*}
where $\mathbf{I}_{nd}$ is the $nd \times nd$ identity matrix.

We assume the parameters $\bTheta_{i}$ for a participant $i$ are generated from participant $i$'s latent variables $\bx_{i}$ by a neural network $\Phi : \mathcal{X} \to \boldsymbol{\Theta}$. 
\begin{align}
    \bTheta_{i} &= \sigma\left(\Phi(\bx_{i})\right) \label{eq:lmc}
\end{align}
Here, the function $\sigma$ is an arbitrary invertible transformation function required to transform the real-valued outputs $\phi(\bx_{i})$ to the domain necessary for the parameters $\bTheta_{i}$. For example, $\btheta_{it}$ might be the probability of a Binomial distribution reflecting the number of trials participant $i$ gets correct on a certain working memory task, in which case the elements of $\phi(\bx_{i})$ corresponding to those parameters must be mapped to the interval $[0, 1]$. For tasks estimating reaction time, $\btheta_{it}$ might be the mean and standard deviation of a log Normal distribution, and the corresponding elements of $\phi(\bx_{i})$ must be mapped to the positive reals.

\paragraph{Full joint model} The joint probability model for the data $\bY$, parameters $\bTheta$, and latent variables $X$ is given by:
\begin{equation}
    p(\bY, \bTheta, X) = \prod_{i=1}^{n}p(\by_{i} \mid \bTheta_{i})p(\bTheta_{i} \mid \bx_{i})p(X) \label{eq:full_joint}
\end{equation}
This model incorporates correlations across cognitive tests in the battery and across individuals in the population. The neural network mixes the \textit{a priori} independent latent variables to introduce correlations in the participant's parameters $\bTheta_{i}$. Furthermore, since we use a common neural network for all participants, this procedure additionally introduces correlations between parameters across participants, as well, e.g., between $\bTheta_{i}$ and $\bTheta_{j}$.

\paragraph{Neural Network (NN) Architecture} The neural network $\phi$ has 2 fully connected hidden layers of size $D$. Each hidden layer is followed by the ReLU activation layer, as in common NN designs \cite{aggarwal2018neural}.  The input layer has size $d$, the latent dimensionality whereas the output layer has size $D$, the number of output parameters. This network transforms a point (latent vector) $x_i$ in a $d$-dimensional space to an output vector of size $D$, which is then transformed using $\sigma(\cdot)$ as shown in \autoref{eq:lmc} to obtain the predicted distributional parameters as illustrated in \autoref{fig:dlvm-architecture}


\subsection{Variational inference}
To train the model described above, we need to perform learning for the latent variables $X$ and the parameters of the neural network $\bw_{\textrm{NN}}$. To do this, we seek the intractable marginal log-likelihood of the data:
\begin{align}
 \log p(Y) &= \log \int p(Y, \bTheta, X)d\bTheta dX\nonumber \\
 &=\log \int p(Y \mid \bTheta)p(\bTheta \mid X)p(X)d\bTheta dX   
\end{align}
To achieve tractability, we introduce a variational distribution $q(X)$. We assume the latent variables factor over the individuals, and introduce Gaussian variational posteriors for the latent variables of each individual:
\begin{align}
    q(X) = \prod_{i=1}^{n}q(\bx_{i}) = \prod_{i=1}^{n} \mathcal{N}(\mathbf{m}_{i}, \mathbf{S}_{i})
\end{align}
As is standard in variational inference, we use this variational distribution to lower bound the marginal likelihood. This enables us to write:
\begin{align}
    \log p(Y) &= \log \int p(\bY, \bTheta, X) d\bTheta \frac{q(X)}{q(X)} dX\nonumber \\
    &= \log \mathbb{E}_{q(X)}\left[\frac{1}{q(X)}\int p(\bY, \bTheta, X) d\bTheta\right] \label{eq:elbo1}
\end{align}

Because the joint distribution factors as in Equation \ref{eq:full_joint}, we can rewrite the inner integral as:
\begin{equation}
    \int p(\bY, \bTheta, X) d\bTheta = \mathbb{E}_{p(\bTheta \mid X)}\left[p(\by \mid \bTheta)\right] \label{eq:joint_as_exps}
\end{equation}

Substituting \autoref{eq:joint_as_exps} into \autoref{eq:elbo1} and applying Jensen's inequality results in the lower bound:
\begin{equation}
    p(Y) \geq \mathbb{E}_{p(\bTheta \mid X)q(X)}\left[\log p(\by \mid \bTheta)\right] - \mathrm{KL}(q(X) || p(X)),
\end{equation}
where $\mathrm{KL}$ represents the Kullback-Lievler Divergence. Exploiting the factorization of the variational distribution $q(X)$ yields:
\begin{align}
    p(Y) \geq \mathbb{E}_{p(\bTheta \mid X)q(X)}\left[\log p(\by \mid \bTheta)\right] - \notag \\ \sum_{i=1}^{n}\mathrm{KL}(q(\mathbf{x}_{i}) || p(\mathbf{x}_{i}))
\end{align}
Finally, again using the factorization of the variational distribution $q(X)$ and the conditional independence assumption between $\by_{ip}$ and $\by_{iq}$ given $\bTheta_{i}$, this evidence lower bound (ELBO) becomes:
\begin{align}
    p(Y) \geq \sum_{i=1}^{n}\sum_{t=1}^{T} \mathbb{E}_{p(\btheta_{it} \mid \bx_{i})q(\bx_{i})}\left[\log p(\by_{it} \mid \btheta_{it})\right] - \notag\\ \sum_{i=1}^{n}\mathrm{KL}(q(\mathbf{x}_{i}) || p(\mathbf{x}_{i}))
    \label{eq:elbo_final}
\end{align}
In each iteration of training, latent variables are sampled from $q(\bx_{i})$ for participant $i$ and used to produce outputs from the neural network $\Phi$. Outputs from this model are transformed using $\sigma(\cdot)$ to the domains required by $\bTheta_{i}$. Given these final samples $\bTheta_{i}$, the log-likelihood of the data $\by_{i}$ is calculated, and then training proceeds by maximizing the ELBO. This procedure can be batched over subsets of participants for large populations.

\subsection{Active learning in distributional LVMs}
In the cognitive testing scenario we study in this paper, standard methods involve serially acquiring data from each test for each participant. Observations are generally assumed to be independent, which leads to inference procedures that are inefficient and limited by the amount of time individuals can sit for tests. Our goal is to determine as efficiently as possible standard cognitive model outputs for individual test takers in order to make repeat testing on different days more practical. In the above distributional LVM model, this problem can be cast as active learning over the \textit{latent variables} $\hat{\bx}$ for each added individual by choosing tests for that individual from which to collect data. To do this, we can compute the mutual information between the outputs of the $t$th test $\hat{\by}_{t}$ and the latent variables $\hat{\bx}$ given a currently collected set of data $\mathcal{D}$. The mutual information is computed using \autoref{eq:mutual_info} where $H[\cdot|\cdot]$ denotes the conditional entropy:
\begin{align}
    I(\hat{\by}_{t} ; \hat{\bx} \mid \mathcal{D}) = H\left[\hat{\by}_{t} \mid \mathcal{D}_{t}\right] - \mathbb{E}_{q(\hat{\bx})}\left[H[\hat{\by}_{t} \mid \hat{\bx}, \mathcal{D}]\right]
    \label{eq:mutual_info}
\end{align}
We estimate both terms above via random sampling. Dropping $\mathcal{D}$ for compactness, the second term can be expanded as:
%
%
\begin{align}
    \mathbb{E}_{q(\hat{\bx})}\left[H[\hat{\by}_{t} \mid \hat{\bx}]\right] &=  \mathbb{E}_{q(\hat{\bx})}\left[\mathbb{E}_{p(\hat{\by}_{t} \mid \hat{\btheta}_{t})p(\hat{\btheta}_{t} \mid \hat{\bx})}\left[-\log p(\hat{\by}_{t} \mid \hat{\btheta}_{t})\right]\right]
\end{align}
enabling the sampling procedure comparable to the one used to compute the ELBO (\ref{eq:elbo_final}). In each iteration of active learning for an individual, we compute the information gain for all cognitive tests in the battery. After choosing a test $t$ (and possibly, for example, a psychometric input $k$) that maximizes information gain, we collect a single sample $y_{it}^{(s)}$ by running a single trial of the chosen cognitive test. This trial is added to the collected data $\hat{\by}$ for this individual, and we then update the latent variables $\hat{\bx}$ with other parameters held fixed through the ELBO. This active learning procedure for a single individual is summarized in Algorithms \autoref{alg:active_learning} and \autoref{alg:update_meu_z}. Algorithm \autoref{alg:active_learning} describes the entire active learning procedure, whereas Algorithm \autoref{alg:update_meu_z} focuses on updating an individual's latent variables given new data.

\begin{algorithm}
\caption{Active Learning with Mutual Information}
\begin{algorithmic}[1]

\STATE Input: \ourmethod{} model \( \Phi \), and data \( \mathcal{D} \)
\STATE $\mu_z, \sigma_z \gets 0, \mathbf{I}$
\STATE \( K \gets \text{number of latent samples} \)
\STATE \( M \gets \text{number of posterior samples} \)

\FOR{ i = 1  to  MAX\_ITERATIONS}
    \STATE Sample \( K \) latent variables \( \hat{x} \sim q(\hat{x} \mid \mu_z, \sigma_z) \)
    \STATE Predict posterior parameters \( \hat{\theta} \sim p(\hat{\mu_z},\hat{\sigma_z} \mid \hat{x}) \)

    \FOR{\( t \) in \( \mathcal{T} \)}
        \STATE Sample \( M \) predictions \( \hat{y}_t \sim p(\hat{y}_t \mid \hat{\mu_z},\hat{\sigma_z}, t) \)
        \STATE Calculate mutual information \( I(\hat{y}_t; \hat{x} \mid \mathcal{D}) \)
    \ENDFOR

    \STATE Select test \( t^* \) with highest \( I(\hat{y}_t; \hat{x} \mid \mathcal{D}) \)
    \STATE Collect new sample \( y^* \) from \( t^* \)
    \STATE Update \( \mathcal{D} \gets \{D  \cup y^*\} \) 
    \STATE Update $N(\mu_z,\sigma_z)$ based on \( D \) using Algorithm \autoref{alg:update_meu_z}
\ENDFOR

\STATE Output: $\mu_z,\sigma_z$

\end{algorithmic}
\label{alg:active_learning}
\end{algorithm}

\begin{algorithm}
\caption{Update Latent Distribution $N(\mu_z,\sigma_z)$}
\begin{algorithmic}[1]
\STATE Input  \(\mu_z,\sigma_z, \Phi\)
\STATE Define prior \( p(\mathbf{X}) \sim \mathcal{N}(\mathbf{0}, \mathbf{I}) \)
\FOR{\(i = 1\) to \(\text{MAX\_ITERATIONS}\)}
    \STATE \text{Sample} \(\mathbf{z}_i \sim N(\mu_z,\sigma_z)\)
    \STATE \text{Compute model output} \( \mathbf{\theta} = \Phi(\mathbf{z}_i) \)
    \STATE \text{Compute log probability} \( L_{\text{data}} = \sum_{t}^{T} \log p(\mathbf{y}_t | \mathbf{\theta}_t,t) \)
    \STATE \text{Compute} \( L_{KLD} = D_{\text{KL}}(q(\mathbf{x} | \mu_z,\sigma_z) \| p(\mathbf{X})) \)
    \STATE \text{Compute loss} \( \mathcal{L} = -L_{\text{data}} + \lambda L_{KLD} \)
    \STATE \text{Backpropagate and update} \( \mu_z,\sigma_z \)
\ENDFOR
\STATE Output: $\mu_z,\sigma_z $
\end{algorithmic}
\label{alg:update_meu_z}
\end{algorithm}


\section{EXPERIMENTS}
\label{sec:experiments}

\subsection{Data collection}

We conducted three separate data collection studies involving three distinct cohorts of young adults. The first study, \train{}, was used for model building. The second was a validation study, following two protocols: \validref{} and \validnew{}. Finally, we conducted a reliability study, \reliab{}, to assess the test-retest reliability of our method. Each study consisted of distinct participant groups. Before joining the study, all participants consented through an IRB-approved form. The studies involved the participants taking a series of cognitive tests via a mobile app. Each participant completed these test batteries in a one-hour session on their respective days of participation, and for their contribution, they received a compensation of \$10. These test sessions were self-conducted: participants downloaded the designated app and were provided with essential troubleshooting guidelines and relevant contact details for any potential issues. In cases where multiple sessions spanned different days, we encouraged participants to undertake them around the same time daily. The test battery encompassed various assessment tasks, including PASAT, Countermanding, Running Span, Numerical Stroop Animals, Magnitude Comparison, Rule Switch Shapes, Simple Corsi, Flanker Arrows, Complex Corsi, Number Line, and Cancellation. Detailed descriptions of these tasks can be found in \cite{pahor2022ucancellation,pahor2022near,rojo_scalable_2023,rojo_accelerating_2023}.

\paragraph{Training protocol \train{}}
This study involved a cohort of 18 participants who underwent 10 sessions of a conventional test battery across 10 consecutive days, resulting in a cumulative 180 sessions of data collection. Conventional test battery tasks included PASAT, which lasted approximately 4 minutes covering numbers 6-15 in about 20 trials; Countermanding, lasting roughly 4 minutes; Running Span with 6 trials at levels 2 and 3, lasting around 5.5 minutes; Numerical Stroop conducted over 2 to 3 minutes;   Simple Corsi for around 5 minutes;  Complex Corsi for 6 minutes; and Cancellation lasting 3 minutes. Participants were afforded short, optional breaks during the sessions. Other tests delivered but not used in our model construction included Flanker, Rule Switch, Number Line, and Magnitude Comparison.

Due to unforeseen technical issues, 79 of these sessions inadvertently omitted observations for some tests and were subsequently excluded from our analysis. The technical issues were corrected as they were discovered to ensure complete testing procedures in the subsequent test sessions. A further subset of 5 sessions displayed extreme outlier performance on one or more tests, such as response timeouts or no correct responses, and were excluded. Following these exclusions, the final dataset consisting of 96 sessions was employed for training the \ourmethod{} model. For this study, each session was treated as an independent collection, and potential correlations within individual participants over different days were not modeled.

\paragraph{Validation protocols \validref{} and \validnew{}}
To compare how \ourmethod{} compares to conventional methods for representing task outputs, we conducted another study in which a new cohort of 33 young adults underwent two testing sessions with at least one day and at most three days between sessions. In one protocol (\validref{}) the participants experienced a conventional procedure identical to \train{}. Task items were delivered sequentially in block form one task at a time in the order shown in \autoref{tab:tb_battery}, for a total of around 280 task items.

In the other protocol (\validnew{}) participants first received a fixed primer sequence of task items distributed across all the tasks. The primer sequence consisted of the following task items, arranged into least divisible units (LDUs): Simple span task, 4 distinct span lengths (4 LDUs), progressing from 4 to 7 items; Complex span task, 4 span lengths (4 LDUs), with 2 spans containing 4 items each and another 2 spans with 5 items each; Countermanding task, 4 trials (1 LDU); Stroop task, 6 trials (1 LDU); PASAT task, 6 trials (1 LDU); and Cancellation task, 2 rows of items (2 LDUs). Therefore, the primer sequence consisted of 26 total task items.

Following the primer sequence, an active data collection sequence commenced, where the most informative next task was selected by Algorithm \ref{alg:active_learning}. The minimum number of trials delivered for any given task is indicated by that task's LDU. These minima were established to mitigate potential task-switching costs because task ordering is variable in the active learning condition. Active learning continued until the total number of task items for a session met or exceeded 100, at which point the session terminated.

Each participant completed the two test batteries in a crossover design with random order. While 33 participants enrolled in the study, only 19 successfully completed both protocols with sufficient data for constructing models. Each of these two different data sources was used to fit two different models.

\begin{table}
    \centering
    \caption{\validref{} Test Battery: Executive function tests delivered sequentially in block form.}
    \label{tab:tb_battery}
    \begin{tabularx}{0.45\textwidth}{|m{0.03\textwidth}|m{0.1\textwidth}|X|}
    \hline 
    Order & Test & Details \\ 
    \hline 
    1 & Simple span & Set sizes 3-8 with two trials per set size \\ 
    \hline 
    2 & Complex span & Set sizes 3-8 with two trials per set size \\ 
    \hline 
    3 & Countermanding & 12 yellow alien trials, 12 blue alien trials, 48 mixed trials \\ 
    \hline 
    4 & Stroop & 60 trials (20 per condition: stimulus size, side, neutral, random) \\ 
    \hline 
    5 & PASAT & 20 trials \\ 
    \hline 
    6 & Cancellation & Dogs for 1 min 10 seconds, cats for 1 min 10 seconds, mixed for 3 min 30 seconds \\ 
    \hline
    \end{tabularx}
\end{table}

Note that Running Span of lengths 2 and 3 were included in the training data (\train{}) for the \ourmethod{} model, but were omitted from further data collection for the validation study. Preliminary analysis revealed that Running Span provided no substantive incremental improvement in modeling working memory performance than the other span tasks alone. Therefore, validation was performed on the subset of tasks found in both the training and the testing datasets.

\paragraph{Reliability protocol \reliab{}}
To establish the test-retest reliability of \ourmethod{}, we ran another study with 8 young adults in which they took a test battery generated by \ourmethod{} twice on different days as part of the \reliab{} protocol. The administered test battery is similar to the \validnew{} protocol described earlier, starting with the same primer sequence followed by an actively generated sequence of tasks governed by the same algorithm.

\subsection{Model construction and training}
We constructed and trained a three-dimensional latent variable model. Given a latent vector \( \mathbf{x}_i = [x_1, x_2, x_3] \), our transformation function in (\autoref{eq:lmc}) maps \( \mathbf{x}_i \) to a 12-dimensional parameter vector \( \mathbf{\bTheta}_i = [\theta_1, \theta_2, \dots, \theta_{12}] \), i.e., \[ f : \mathbb{R}^3 \rightarrow \mathbb{R}^{12}, \] where $f$ is the transformation function. Put simply, the neural network $\phi$ takes in $d = 3$ and outputs $D =12$.


The resulting vector \( \mathbf{\bTheta} \) represents predicted parameters for the different distributions we used to model the cognitive tests. Out of the numerous cognitive task data we gathered, our analysis concentrated on seven tasks, grouped by data type into three primary categories: Timing Tasks, consisting of Stroop and Countermanding, where we assessed reaction times; Psychometric Tasks, encompassing both Simple and Complex spans; and Accuracy Tasks, including Cancellation, PASAT, Running Span with Length 2, and Running Span with Length 3. For the purposes of this study, data from all other tests were excluded. We chose different likelihood distribution types for the different cognitive tests to model the item-level observations.

\paragraph{Timing tasks} For stroop and countermanding tasks, we employed log-normal distributions to model the reaction times. These are described using two parameters: the mean (denoted as $\mu_\tau$) and the standard deviation (represented by $\sigma_\tau$).

\paragraph{Span tasks} Both Simple and Complex spantasks were depicted using a psychometric sigmoid curve. This curve is characterized by two parameters: the threshold (noted as $\theta_\psi$) and the spread (expressed as $\sigma_\psi$).

\paragraph{Accuracy tasks} The tasks centered on accuracy were captured using binary distributions. These distributions rely on a single parameter, the probability of success (indicated by $p$). The accuracy tasks include Cancellation, PASAT, Running Span with length 2, and Running Span with length 3.

This formulation corresponds to 4 parameters from timing tasks, 4 parameters from span (complete \& simple) tasks, and 4 parameters from accuracy tasks, making a total of 12 output parameters.

As with the ELBO in \autoref{eq:elbo_final}, the \ourmethod{} model was trained to minimize the loss in \autoref{eq:loss_fn} using the \train{} dataset.:

\begin{align}
    L(X) = -\sum_{i=1}^{n}\sum_{t=1}^{T} \log p(\by_{it} \mid \btheta_{it}) + \notag\\ \lambda \sum_{i=1}^{n}\mathrm{KL}(q(\mathbf{x}_{i}) || p(\mathbf{x}_{i}))
    \label{eq:loss_fn}
\end{align}

 As the model adjusted the positions of individual data points in the latent space, the aim was to optimize the transformed observations to achieve the highest possible log probability on the training data/observations. Additionally, we incorporated a regularization term weighted by a hyperparameter $\lambda$ to constrain the latent space and prevent overfitting, ensuring that our model provided a robust and generalized data representation. The results presented in this paper were obtained by training \ourmethod{} on the full \train{} dataset (n = 96), following a thorough process of hyperparameter selection. The final model was trained with a regularization parameter of $\lambda = 0.1$ for up to 8,000 iterations in full-batch mode, using the Adam optimizer with a learning rate of $0.001$.

Prior to training the final model, preliminary experiments were conducted using a split dataset, with 90 samples in a training set and 6 samples in a validation set. Various hyperparameter values were explored, including dimensionalities (12, 6, 2, 3), regularization parameters ($\lambda = 10, 1, 0.1, 0.01, 0.001$), and learning rates ($lr = 0.1, 0.01, 0.001, 0.0001, 0.00001$). The final hyperparameters were selected based on performance on the validation set and subsequently used for training the complete model from \train{} data.


\subsection{Latent position update}
For each iteration of active learning, the latent distribution estimated for an individual was computed using Algorithm \autoref{alg:update_meu_z}. Each update was run for a maximum of 4,000 iterations in a full-batch mode using the Adam optimizer with a learning rate of 0.001. The latent distribution was initialized as \(N(\mu_z, \sigma_z) \sim \mathcal{N}(\mathbf{0}, \mathbf{I})\), which means it started with a mean position at the origin (0, 0, 0) and a unit \( l_2 \)-norm around it. The hyperparameters \(M\) and \(K\) presented in Algorithm \autoref{alg:active_learning} were assigned values of 500 and 1000, respectively, chosen based on the performance the validation set.

\subsection{Evaluation}
To assess the model's performance on new data, we utilized datasets from the \validref{}, \validnew{}, and \reliab{} protocols. Because these datasets were gathered from a distinct group of participants from the \train{} training set, this approach offers an out-of-sample evaluation of \ourmethod{}'s effectiveness.  

\paragraph{Validation}
We employed the \validnew{} data to estimate each task output using the trained \ourmethod{} model. This model discerned the interrelationships between all performance variables of the test battery for a reference (training) group, condensing them into a concise latent variable representation. This data evaluation method aligns with the Bayesian approach and is termed \textit{Distributional Active Learning} (\ouractivemethod{}). It indicates that new task items can be optimally chosen in real-time by evaluating the uncertainties spanning all output variables of interest. One of the advantages of \ourmethod{} is the ability to select task items optimally, although any collection of task output data could be used to estimate latent variable positions and corresponding distribution parameters.

On the other hand, the \validref{} protocol data were utilized to construct models for each task output based solely on that specific task's data. This method represents the traditional procedure for assessing scientific data under the frequentist paradigm, which we have denoted as \textit{Independent Maximum Likelihood Estimation} (\oldmethod{}). Consequently, all conventional models in the subsequent analysis carry the label \modelref{}, while our new modeling framework is designated \modelnew{}. The main hypothesis of this research can therefore be stated as \modelnew{} results are expected to be congruent with \modelref{} results, albeit with greater flexibility in selecting task items in order to generate a more efficient testing procedure. To evaluate this hypothesis, we performed equivalence testing using the Two One-Sided T-test (TOST) to determine the tolerance limits of the final estimates from both \modelref{} and \modelnew{} at a significance level of ( p = 0.05 ). 

\paragraph{Test-retest reliability} To assess the reliability of the estimates produced by \ourmethod{}, a separate cohort of participants in the \reliab{} cohort undertook the active learning procedure on two separate days, resulting in repeated \modelnew{} models. On each occasion, the model generated parameter estimates for the individual participants. We subsequently conducted a paired correlation analysis on these estimates to determine the test-retest reliability of \ourmethod{} given 100 total task items, 74 of which were actively selected.
\section{RESULTS}
\label{sec:results}

\begin{figure*}[!htb]
    \centering
    \includegraphics[width=\textwidth]{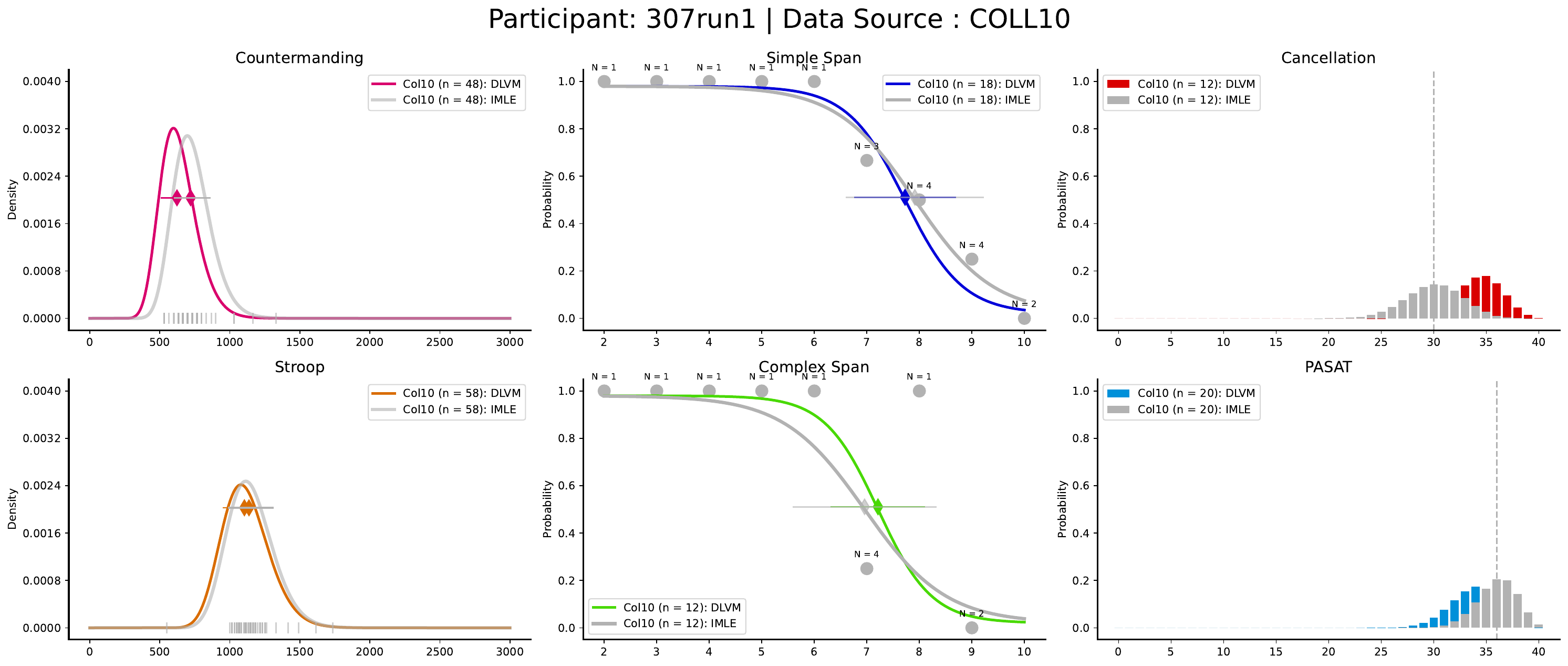}
    \caption{Median Fit of the trained \ourmethod{} model to the conventionally acquired \train{} data (colored) compared with the \oldmethod{} fit to data (gray). The median fit was identified based on the log probability of the data given the predicted parameters by the \ourmethod{} model. Timing outputs in milliseconds (left column) were fit with log-normal distributions. Item response times are indicated by short vertical lines, means by diamonds, and standard deviations by horizontal lines. Psychometric outputs (middle column) were fit by logistic sigmoids as a function of items to be remembered. Proportion correct for test times is indicated by circles, thresholds by diamonds and spreads by horizontal lines. Accurate completion of test items (right column) is fit by binomial distributions reflecting a hypothetical test containing 40 items. The proportion of test items successfully completed is indicated by vertical dashed lines.}
    \label{fig:in-sample_fits}
    \vskip -0.1in
\end{figure*}
In this section, we describe our results and demonstrate a \ourmethod{} model that successfully learned the correlations within the \train{} data and performs well on unseen data. We compare our method against \oldmethod{}. Estimates generated by our method are labeled with \ourmethod{} whereas those generated by the traditional method are labeled with \oldmethod{}. 

\subsection{Distribution fits for training Data}
The trained \ourmethod{} model was able to capture the intra-participant and inter-participant correlations in the cognitive tests by projecting the observation (feature-space) performance to unique positions in the 3-dimensional latent space. 

\autoref{fig:in-sample_fits} shows the median in-sample fit of the trained model across six different tasks. The \ourmethod{} fits are similar to the \oldmethod{} fits in this case, indicating that the trained model has learned a reasonable latent variable representation of the corresponding feature space of task outputs. The binomial distributions reflecting accuracy (i.e., PASAT and Cancellation) appear to be the least well aligned in this example, as well as the overall dataset. This observation can be at least partly attributable to the nature of the binomial distribution, in which central dispersion values are directly related to the total number of data points. The resolution of distributions reflecting accuracy tasks is therefore fundamentally limited by the amount of data available to fit them. The ability of \ourmethod{} to fit distributions using outside knowledge could represent an attractive alternative to the standard approach of accruing enough single-task data sufficiently removed from floor and ceiling effects to yield an interpretable accuracy score. 

\subsection{Distribution fits for testing data}
\begin{figure*}[!htb]
    \centering
    \includegraphics[width=\textwidth]{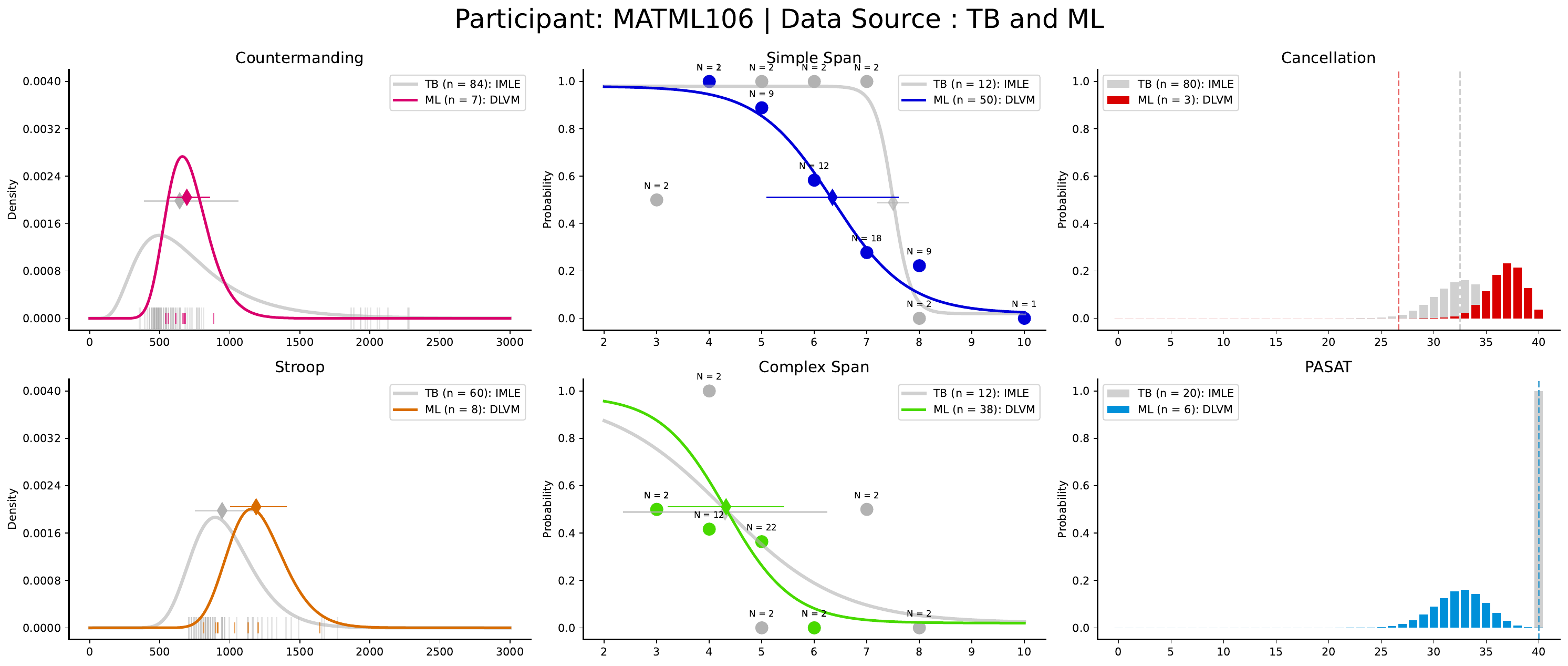}
    \caption{Distributions fitting test battery outputs for conventional test battery data (\modelref{}, gray) and for machine learning test battery data (\modelnew{}, colors) for a median-fit participant (ID: MATML106). The median is based on the root-mean-square error (RMSE) between the \oldmethod{} and \ourmethod{} parameters. Plotting conventions as in \autoref{fig:in-sample_fits}.}
    \label{fig:out-of-sample_fits}
    \vskip -0.1in
\end{figure*}

\autoref{fig:out-of-sample_fits} shows modeled test outputs for the median out-of-sample fit for both \modelref{} (gray) and \modelnew{} (colors). Because \modelref{} by definition can only use the data visible on each plot axis to make predictions, the models and data would be expected to visibly correspond. That generally appears to be the case here. \modelnew{} is free of this constraint, however, and can incorporate other information deemed informative. In this case, the extra information incorporated into the model is how people generally perform on tests of this sort, as well as this individual’s performance on all the other tests in the battery.

The noticeable discrepancies in this case are worthy of additional inspection. Countermanding shows a number of long response times of around 2 seconds in the \modelref{} run, leading to a bimodal distribution. The effect of these long times in widening the lognormal distribution and raising its mean can be observed. Because the conventional single-task frequentist modeling framework does not allow outside information to be incorporated into the model, altering the representation should would require labeling some data as ``outliers'' and excluding them from further analysis. Alternatively, one could collect larger amounts of data to empirically overwhelm outliers with valid data. The former strategy has the potential to increase estimator bias, while the latter strategy has the potential to increase estimator variance. Both strategies may ultimately reduce overall estimator error if applied properly. One of the main advantages of \ourmethod{}, however, is its ability manage the bias/variance tradeoff empirically, by allowing all data across all sessions and tasks to contribute to all distribution fits, as well as allowing an inductive bias via a Bayesian prior belief.

In this example, the actively acquired Countermanding data exhibited no obvious outliers. This might have occurred by chance or from the smaller overall number of active samples for this task or from the task-switching nature of \ouractivemethod{}. In any case, the means of the two Countermanding distributions are concordant. A greater disparity is apparent for the Stroop means, but both model Stroop distributions appear generally reflective of their underlying data. A similar observation can be made for the Span models.

Greater discrepancies are apparent for the binomial models of task accuracy, however, as was the case for the in-sample example. This behavior is noticeable across the dataset and could result from a multitude of factors. As noted above, tasks whose primary output is accuracy may be less well suited to \ourmethod{}s generally, at least for the limited number of test items typically employed for such tests. Further research may be able to discern the relative utility of different tasks for constructing the latent variable models
\begin{figure}[!hb]
    \centering
    \includegraphics[width=0.48\textwidth]{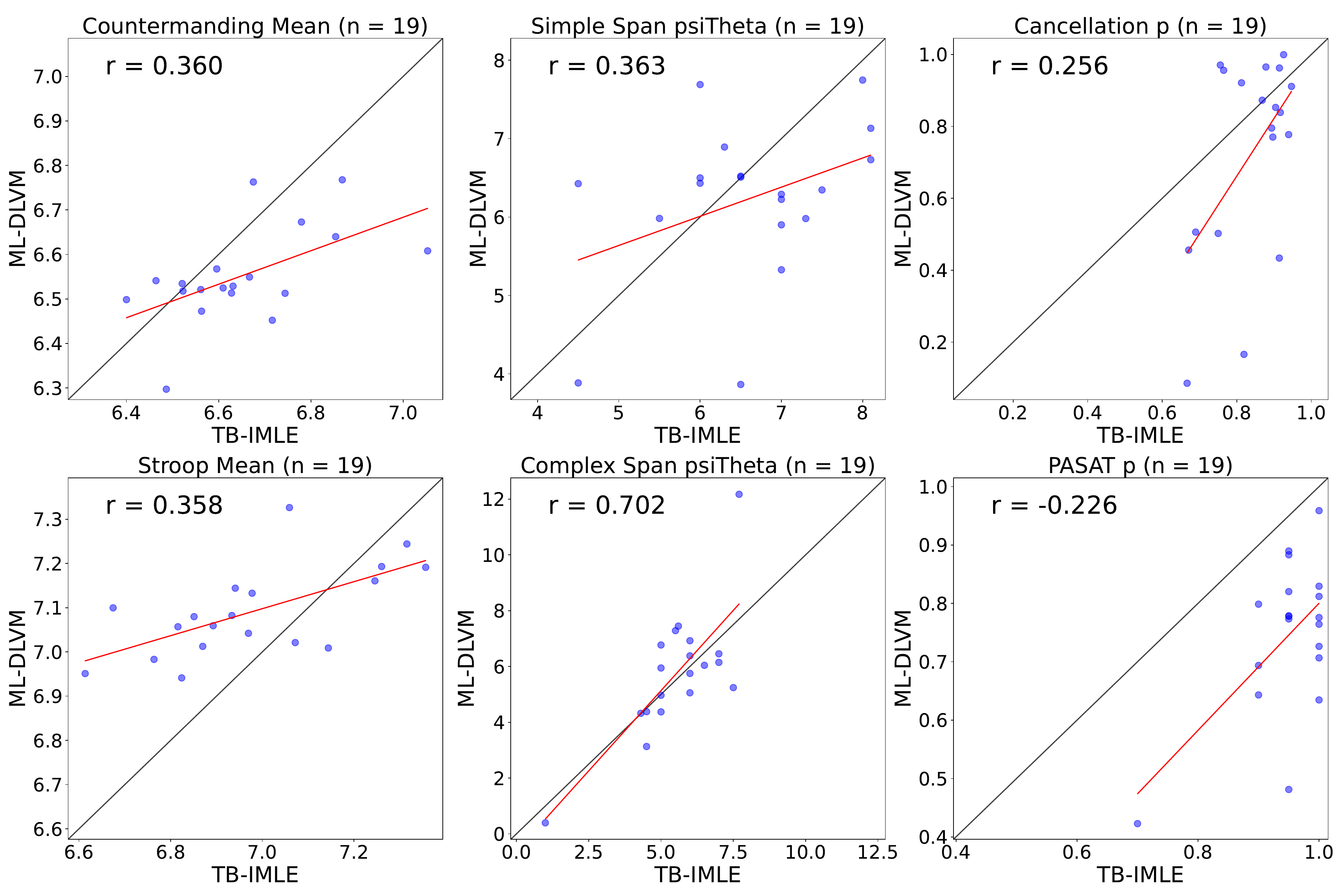}
    \caption{Similarity of primary test outputs (i.e., means and thresholds) between the conventional test battery + independent maximum likelihood estimation (\modelref{}) and machine learning battery + distributional latent variable modeling (\modelnew{}). Timing task mean responses (left column) are in units of log milliseconds, psychometric task thresholds (center column) are in units of items recalled, and accuracy tasks (right column) are in units of probability correct. Numerical values are intra-class correlation coefficients.}
    \label{fig:validation}
    \vskip -0.1in
\end{figure}
\subsection{Similarity between \modelref{} and \modelnew{}}
The correspondence between the two testing procedures is quantified in \autoref{fig:validation}. As judged by the indicated correlation coefficients, correspondence is better for some tests than others. The Simple Span threshold has a correlation coefficient of $0.363$, for example, while the Complex Span threshold exhibited the highest correlation coefficient of $0.702$. Given that these two tasks are quite similar, the disparities in correlations across tasks do not seem to be attributable solely to task differences. We also see that the accuracy tests have low correspondence with PASAT, exhibiting a negative intra-class correlation. This could be attributed to the ceiling effects of these accuracy tests that affect the \modelref{} estimates whereas \ourmethod{} is able to overcome them by leveraging data from other tests.

Equivalence tests using the Two One-Sided T-tests (TOST) approach demonstrated that the two methods are statistically equivalent within calculated tolerance limits at a  \( p = 0.05 \) significance. For reaction time tasks, modeled as log-normal distributions, the methods are equivalent with tolerance values below 0.2. In span tasks, the methods are equivalent with tolerances of 0.92 for simple span and 1.27 for complex span. The tolerance values for equivalence are larger for accuracy tasks, with tolerances of 0.26 for PASAT and 0.23 for the Cancellation task. These tolerances are acceptable for the reaction time tasks and span tasks but are relatively loose for accuracy tasks that suffer from ceiling effects, more so for the TB-IMLE estimates. 
\begin{figure}[!ht]
    \centering
    \includegraphics[width=0.48\textwidth]{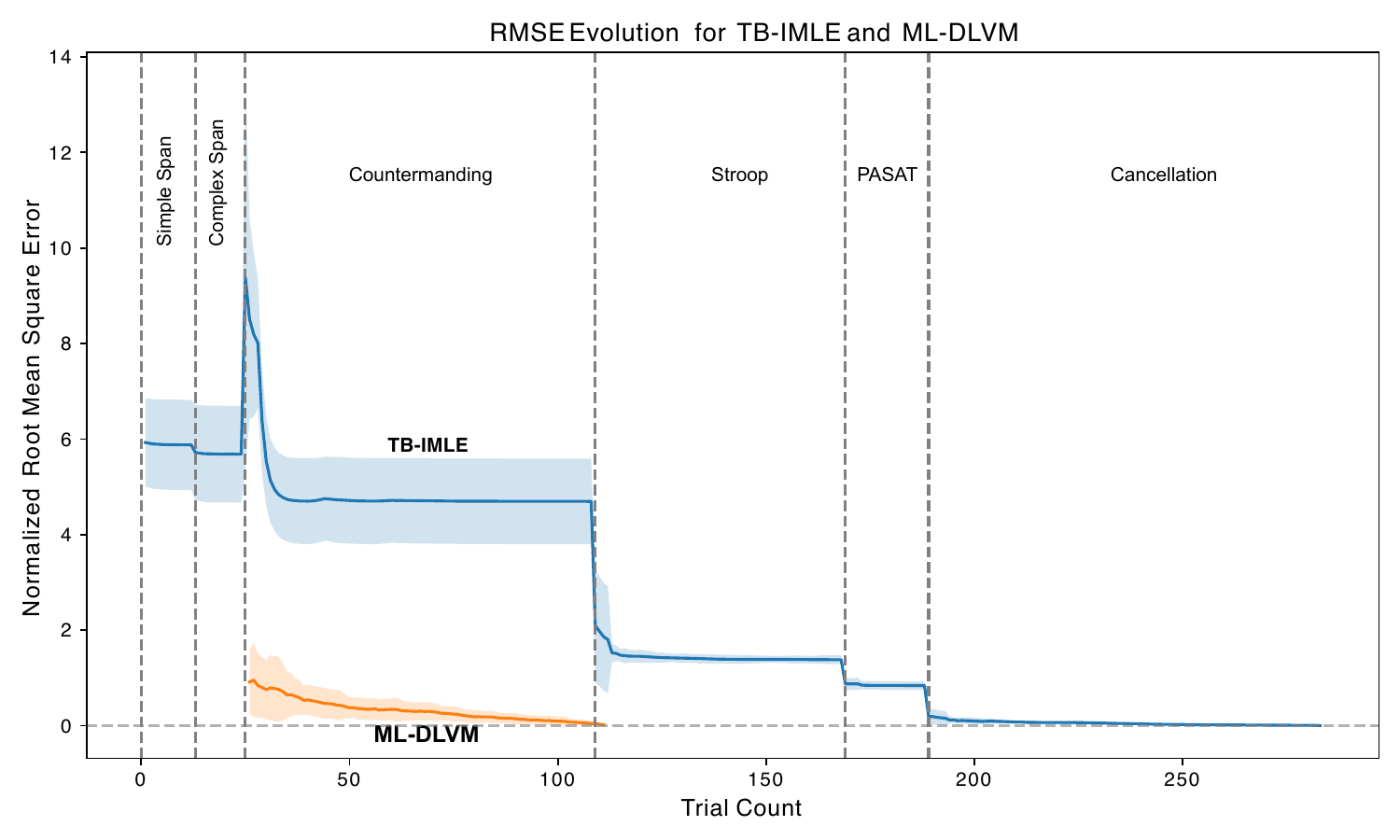}
    \caption{Comparison of average convergence rates toward final estimates for the conventional test battery + independent maximum likelihood estimation (\modelref{}, blue) versus the machine learning battery with distributional latent variable modeling (\modelnew{}, orange). Values for the latter are only shown following primer sequence completion. Root-mean-square error values compared to final estimates were computed as the proportion of the total range of each test output and summed. Lines are means across the cohort of 19 individuals while shaded areas represent ±1 standard deviation.}
    \label{fig:efficiency}
    \vskip -0.1in
\end{figure}
\subsection{Efficiency}
The efficiency of \ourmethod{} compared to the conventional test battery can be seen in \autoref{fig:efficiency}. Substantial redundancy exists for the conventional test battery, with many tests proceeding long after the central tendency of the underlying distribution has been determined reasonably well. Note that \ourmethod{} is quantified only after it has completed its 26-sample primer sequence. The effect of this primer sequence appears to be able to place the latent variable position very close to its final position with 26 fixed test items. The modeling framework provides this efficiency boost, having learned from previous data how best to interpret a relatively small amount of incremental data from a new individual. Active learning provides the second boost, systematically probing the latent variable space for better representations by determining the best task item to deliver next.

\subsection{Test-retest reliability}
\begin{figure}[!ht]
    \centering
    \includegraphics[width=0.48\textwidth]{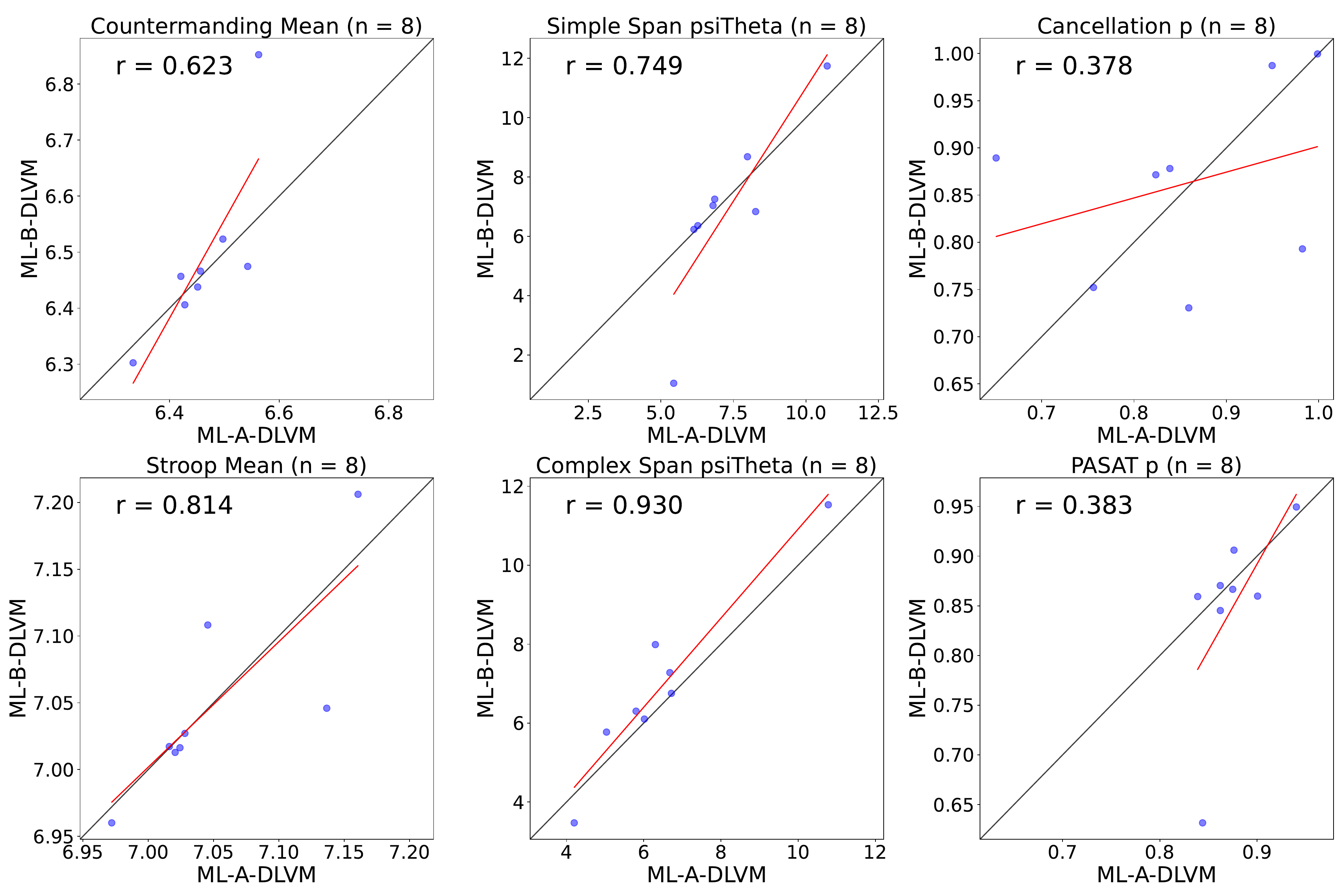}
    \caption{Similarity of primary test outputs between repeats of \modelnew{} on different days (A versus B). Plotting conventions as in \autoref{fig:validation}}
    \label{fig:reliability}
    \vskip -0.1in
\end{figure}

Correspondence between repeats of the machine learning test battery within the same individual on different days for a different cohort of young adults is shown in \autoref{fig:reliability}. Test-retest reliability overall is high, indicating both the robustness of the underlying executive function constructs over time and the estimation consistency. This outcome is particularly encouraging regarding this modeling approach given that the set of task items each day could be completely different. Further, \ourmethod{} is not specifically designed for test-retest reliability \textit{per se}, only uncertainty minimization, making this result even more encouraging that the underlying latent structure is being adequately captured.
Tasks with accuracy outputs (i.e., Cancellation and PASAT) give the poorest reliability. It is possible that these tasks are not particularly helpful at reducing uncertainty in other test results, either because they are inherently unreliable or poorly correlated with the other results or possibly both. One of the ancillary benefits of these machine learning methods is the ability to quantify the overall utility of very different testing procedures for their informativeness in representing cognitive performance. Further application of these methods may be able to identify absolute utility for different task categories to yield a more informative future test battery composition.

\section{DISCUSSION AND FUTURE WORK}
\label{sec:discussion}

\ourmethod{} models evaluated using \ouractivemethod{} achieve our goal of accommodating individual variation in cognitive test performance with accurate individualized cognitive models by leveraging both intrasubject and intersubject correlations at the item level. By harnessing these correlations, we can minimize the amount of data required to estimate cognitive performance across all tests. This conclusion is supported by our results where we showed that we provide accurate item-level predictions for each task and individual with fewer task items than is customarily delivered today. Even when trained with only the 26 primer data points, \modelnew{} is considerably better trained than is \modelref{} with many more data points, as illustrated by \autoref{fig:efficiency}.

\ourmethod{} employs a combined loss function to integrate information from multiple cognitive tasks, each modeled by distinct distributional types. By comparing it to the traditional \modelref{} approach we see nuanced differences in the estimates across tasks. We found stronger agreements between the two approaches from non-accuracy tasks, for example. Generally, \ourmethod{} shows lower variance in estimates and suffers less from ceiling effects associated with accuracy tasks. The ability to draw information from other tasks when making inferences about a given task removes it from the data-driven ceiling/flow effects and leads to more confident estimates. Furthermore, this cross-information makes \ourmethod{} more robust to outliers and missing or limited data or a given test.  

While this modeling framework makes effective use of individual data, it is also scalable to take advantage of population data when available, including much larger populations than demonstrated in this paper. Collectively, these achievements advance our long-term goal of providing rapid, equitable cognitive assessments for young students in order to design optimal math lessons for them on any given day. The substantially reduced testing times mean that testing for a few minutes at the beginning of class each day may be sufficient to obtain actionable inferences for daily individualized lesson planning.

Challenges and further work remain. The most substantial uncertainty is the overall accuracy of \modelnew{} as implemented. Generally speaking, primary test outputs for \modelnew{} at $\sim$100 samples compared against \modelref{} at $\sim$280 samples with correlation coefficients of around 0.3–0.7 for a young adult population. Whether this value is above a target quality threshold depends on the particular application. It is possible that lower reliability of \modelref{} in this cohort reduces the correlation values. The lower test-retest reliability values typically yielded by \oldmethod{} procedures, in the range of 0.4 or lower \cite{rouder2023correlations}, when compared against \ourmethod{} reliabilities of 0.6 to 0.9, are consistent with this interpretation. In any case, \ourmethod{} provides more flexibility for designing cognitive test batteries out of new task designs that may employ unusual output measures.


For the first time, we have constructed nonlinear item-level latent-variable predictive models of intra-individual performance, complete with associated uncertainties. We have intentionally focused on evaluating predictive validity using individual test battery outputs, but other options are more attractive in the long run. In particular, the nature of the learned latent variable space itself should be reflective of relevant cognitive constructs. Future work will explore these latent variable spaces directly. Finally, we aim to make our method general and apply it to many applications suitable for latent variable modeling of distributional data.
\section*{Acknowledgment}

The research reported here was supported by the EF+Math Program of the Advanced Education Research and Development Program (AERDF) through funds provided to Washington University in St. Louis. The opinions expressed are those of the authors and do not represent views of the EF+Math Program or AERDF.

\ifCLASSOPTIONcaptionsoff
  \newpage
\fi

\bibliographystyle{IEEEtran}
\bibliography{11_references}

\end{document}